\documentclass{article}
\newif\ifarxiv\arxivtrue 
\ifarxiv%
\else%
  \usepackage{neurips_2021_ml4ps}
\fi
\newcommand{\der}[2]{\frac{\mathrm{d} #1 }{\mathrm{d} #2 }}

\usepackage{wrapfig}
\usepackage{caption}
\usepackage{subcaption}
\usepackage{booktabs}
\usepackage{float}
\usepackage[utf8]{inputenc}
\usepackage{amsmath}
\usepackage{amsthm}
\usepackage{amssymb}
\usepackage{amsfonts}
\usepackage{mathtools}
\usepackage{hyperref}
\usepackage{bm}
\usepackage{comment}
\date{}
\ifarxiv%
  \makeatletter
  \renewcommand\section{\@startsection {section}{1}{\z@}%
    {-3.25ex \@plus -1ex \@minus -.2ex}%
    {1.5ex \@plus .2ex}%
    {\raggedright\normalfont\Large\bfseries}}
  \makeatother
  \renewcommand{\small}{\normalsize} 
\else
  \renewcommand{\paragraph}[1]{\par\textbf{#1}}
\fi
\sloppypar

\usepackage[framemethod=tikz]{mdframed}
\usetikzlibrary{shadows}
\definecolor{captiongray}{HTML}{555555}
\ifarxiv
\mdfsetup{%
  innertopmargin=1.0ex,
  innerbottommargin=1.0ex,
  innerleftmargin=0.5em,
  innerrightmargin=0.5em,
  linecolor=captiongray,
  linewidth=0.5pt,
  roundcorner=1pt,
  shadow=false
}
\else
\mdfsetup{%
  innertopmargin=1.0ex,
  innerbottommargin=0.0ex,
  innerleftmargin=0.5em,
  innerrightmargin=0.5em,
  linecolor=captiongray,
  linewidth=0.5pt,
  roundcorner=1pt,
  shadow=false
}
\fi
\newtheoremstyle{Hogg}
  {0.5\baselineskip}
  {0.5\baselineskip}
  {}
  {}
  {\itshape}
  {:}
  {1ex}
  {}
\theoremstyle{Hogg}

\usepackage{array}
\newcolumntype{C}[1]{>{\centering\arraybackslash}m{#1}}
\newcolumntype{L}[1]{m{#1}}
\usepackage{multicol}
\usepackage{multirow}
\renewcommand{\v}{\mathbf{v}}
\newcommand{\q}{\mathbf{q}}
\newcommand{\p}{\mathbf{p}}

\ifarxiv
\else

\fi

\title{\bfseries%
A simple equivariant machine learning method for dynamics based on scalars}
{
\author{%
Weichi~Yao\footnote{Department of Technology, Operations, and Statistics, Stern School of Business, New York University},~~
Kate~Storey-Fisher\footnotemark[4],~~
David~W.~Hogg\footnote{Flatiron Institute, a division of the Simons Foundation},\,\footnote{Center for Cosmology and Particle Physics, Department of Physics, New York University}~~
\& Soledad~Villar\footnote{Department of Applied Mathematics and Statistics, Johns Hopkins University}\;\footnote{Mathematical Institute for Data Science, Johns Hopkins University}%
}}

\renewcommand{\ldots}{.\,.\,}

\begin{document}
\maketitle

\begin{abstract}\noindent 
Physical systems obey strict symmetry principles. We expect that machine learning methods that intrinsically respect these symmetries should have higher prediction accuracy and better generalization in prediction of physical dynamics. In this work we implement a principled model based on invariant scalars \cite{villar2021scalars}, and release open-source code. We apply this \textsl{Scalars} method to a simple chaotic dynamical system, the springy double pendulum. We show that the Scalars method outperforms state-of-the-art approaches for learning the properties of physical systems with symmetries, both in terms of accuracy and speed. Because the method incorporates the fundamental symmetries, we expect it to generalize to different settings, such as changes in the force laws in the system.
\end{abstract}

\section{Introduction} 
\begin{wrapfigure}{r}{0.22\textwidth} 
    \centering
    \includegraphics[scale=0.1]{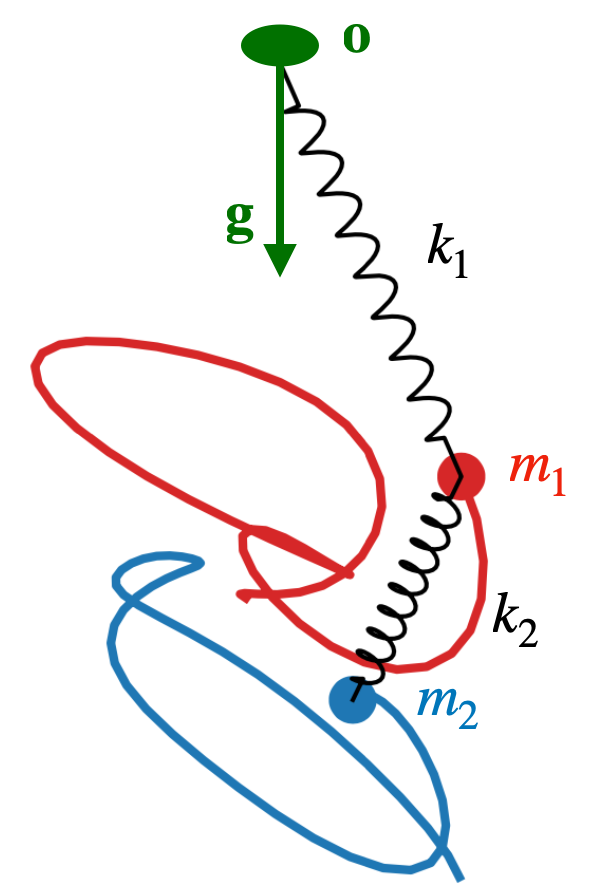}
    \caption{ 
    \small The springy double pendulum used in this work.}
    \label{fig:system}
\end{wrapfigure}

The laws of physics are coordinate-free and equivariant with respect to rotations, translations, and other kinds of transformations. Loosely speaking, a model is equivariant with respect to a group $G$ (such as the rotation group, for example) if the transformation of all the inputs by a group element $g$ leads to an output that is transformed by that same group element. A model is coordinate-free if a change of coordinates applied to all inputs leads to an output that has that same change of coordinates applied.  
Enforcing fundamental symmetries is crucial in the physical sciences. For instance, encoding rotational symmetries significantly improves predictions in molecular dynamics 
\cite{anderson2019cormorant, wang2021molecular, schutt2021equivariant};
quantum energy levels on lattices obey symmetries that should be enforced in analysis tasks \cite{choo2018quantum, vieijra2020quantum, luo2021quantum}; many problems in cosmology also rely on spatial symmetries, such as the translational and rotational invariance of weak lensing maps \cite{Yoo_2018, Gupta_2018}; and it has been noticed that enforcing relevant symmetries can improve turbulent flow prediction accuracy compared to the generic machine-learning methods~\cite{Ling2016turbulence,Chidester2018turbulence,Mattheakis2019turbulence} that do not naturally incorporate symmetries but only gradually improve the approximation. 

There are different ways to enforce symmetries in machine learning models in general \cite{cohen2019gauge, weiler2021coordinate}. In the simplest case---if we have enough training data---a sufficiently flexible model will naturally learn symmetric laws when the prediction task is in fact symmetric. If there are no enough data to learn the symmetries, one could consider adding training data rotated and translated (say) copies of training data elements; this is known as data augmentation \cite{huertas-company2015galaxy, aniyan2017radio, dominguez-sanchez2018galaxy, gonzalez2018augment}. 
Another path for the enforcement of symmetries is to consider the composition of linear equivariant layers on tensor inputs with non-linear compatible activation functions; this typically requires irreducible representations~\cite{kondor2018n, thomas2018tensor, fuchs2020se} and code implementations are available for some examples but not for general symmetries. Recent work by \cite{Wang2021dynamics} builds on the methods of~\cite{Worrall2019scale,Weiler2019E2rotation,Cohen2016Conv} and enforces symmetries (such as rotation and scale) on convolutional neural networks to improve the prediction of physical dynamics. Another approach is to parametrize the space of equivariant linear functions by writing the corresponding constraints that the equivariance enforces, and then solve a large linear system \cite{finzi2021practical}.  

Recently, a new approach to enforcing physical symmetries in machine learning models was introduced in \cite{villar2021scalars} that is simple and powerful compared to existing approaches. It proposes to construct explicitly invariant features (scalars) and use these as inputs to the model, in a way that is universally approximating. The method connects to the way classical physicists represent physical law in terms of coordinate-free scalar, vector, and tensor forms. However, the scalars approach has no publicly available implementation yet. In this work, we implement this method and apply it to learning the dynamics of a springy double pendulum, which is a standard problem in this area. We show that the scalars approach achieves state-of-the-art accuracy for this problem while being considerably faster and simpler than other approaches.

\section{Modeling invariant and equivariant functions with scalars} \label{sec:scalars}

In this work we focus on learning functions that are invariant or equivariant with respect to the orthogonal group $\mathrm{O}(d) = \{\mathbf R\in \mathbb {R}^{d\times d}: \mathbf R^\top \mathbf R =I \}$. The fundamental theorem of invariant functions for $\mathrm{O}(d)$ states that a function $\mathtt f:(\mathbb R^d)^n \to \mathbb R$ is $\mathrm{O}(d)$-invariant (i.e. $\mathtt f(\mathbf R \v_1,\ldots, \mathbf R \v_n) = \mathtt f(\v_1, \ldots, \v_n)$ for all $\mathbf R \in \mathrm{O}(d)$ and all $\v_i \in \mathbb R^d$) if and only if $\mathtt f$ can be expressed as a function of the scalar products of the input vectors:
\begin{align}
    \mathtt f(\v_1,\ldots, \v_n) = \tilde {\mathtt f }\left( (\v_i^\top \v_j)_{i,j=1}^n \right).\label{eq:scalar_output}
\end{align}

The formulation in \cite{villar2021scalars} extends this characterization to equivariant vector functions, stating that a function $\mathtt h:(\mathbb R^d)^n \to \mathbb R^d$ is $\mathrm{O}(d)$-equivariant (i.e. $\mathtt h(\mathbf R \v_1,\ldots,\mathbf R \v_n) = \mathbf R \mathtt h(\v_1,\ldots,\v_n)$ for all $\mathbf R$ and $\v_i$) if and only if $\mathtt h$ can be expressed as a linear combination of the input vectors, where the coefficients are scalar invariant functions of the inputs:
\begin{align}
    \mathtt h(\v_1,\ldots, \v_n) = \sum_{s=1}^n \tilde {\mathtt f}_s \left( (\v_i^\top \v_j)_{i,j=1}^n \right)\v_s.\label{eq:vector_output}
\end{align}

We construct $\mathrm{O}(d)$-invariant (equivariant) neural networks using multi-layer perceptrons (MLPs) that take the scalar products of the input vectors as input features. These neural networks are invariant (equivariant) by construction, and can universally approximate any invariant (equivariant) function since MLPs are universal.

\section{Application: Springy double pendulum}

We consider the dissipationless spherical double pendulum with springs, with a pivot $o$ and two masses connected by springs (Figure~\ref{fig:system}). The kinetic energy $\mathcal{T}$ and potential energy $\mathcal{U}$ of the system are given by
\begin{align}
    \mathcal{T} =&\;\frac{|\mathbf{p}_1|^2}{2m_1} +\frac{|\mathbf{p}_2|^2}{2m_2}, \label{eq:energy_T}\\
    \mathcal{U} =&\;\frac12 k_1(|\mathbf{q}_1-\mathbf{q}_o|-l_1)^2 + \frac12 k_2(|\mathbf{q}_2-\mathbf{q}_1|-l_2)^2 
    -m_1\,\mathbf{g}\cdot (\mathbf{q}_1-\mathbf{q}_o)- m_2 \,\mathbf{g}\cdot  (\mathbf{q}_2-\mathbf{q}_o), \label{eq:energy_U}
\end{align}
where $\mathbf{q}_1, \mathbf{p}_1$ are the position and momentum vectors for mass $m_1$, similarly $\mathbf{q}_2, \mathbf{p}_2$ for mass $m_2$, and a position $\mathbf{q}_o$ for the pivot. The springs have scalar spring constants $k_1$, $k_2$, and natural lengths $l_1$, $l_2$. The gravitational acceleration vector is $\mathbf{g}$. 

We consider the task of learning the dynamics of the pendulum; the goal is to predict its trajectory at later times
from different initial states.
For this task, the parameters of the model $m_1, m_2, k_1, k_2, l_1, l_2$ are fixed during training and test but unknown. 
Our training inputs are $N$ different initializations at $t_0=0$ of the pendulum positions and momenta, and the labels are the positions and momenta at a set of $T$ later times $t$:
\begin{equation}
\mathbf{z}^i(t)=(\mathbf{q}^i_1(t),\mathbf{q}^i_2(t),\mathbf{p}^i_1(t),\mathbf{p}^i_2(t)), \quad t\in\{t_0=0, t_1,\ldots,t_T\} \text{ and } i \in\{1,\ldots, N\}. \label{eq.training}
\end{equation}
To this end we aim to learn the function $f$ that predicts the dynamics: 
\begin{equation}
\begin{aligned}
    f: (\mathbb R^3)^4\times \mathbb{R}^3 \times \mathbb{R}^3 \times \mathbb R &\to (\mathbb R^{3})^4  \\
    (\mathbf{z}(0),\mathbf{q}_o,\mathbf{g},\Delta t) &\mapsto  \mathbf{\hat z}(\Delta t).
\end{aligned}\label{eq:goal_F}
\end{equation} 
This function is $\mathrm{O}(3)$-equivariant in both position and momentum, and translation equivariant in position, namely, for all $\mathbf{R}\in \mathrm{O}(3)$ and all $\mathbf{w}\in \mathbb{R}^3$ we have
\begin{align} 
f( ( \mathbf{R}\mathbf{q}_{1} +\mathbf{w} , \mathbf{R}\mathbf{q}_{2} +\mathbf{w}, \mathbf{R} \mathbf{p}_{1}, \mathbf{R}\mathbf{p}_{2}),\mathbf{R}\mathbf{q}_o+\mathbf{w},\mathbf{R}\mathbf{g}) = \mathbf{R} f( (\mathbf{q}_{1},\mathbf{q}_{2} ,\mathbf{p}_{1} ,\mathbf{p}_{2}),\mathbf{q}_o, \mathbf{g}) +\mathbf{w} .
\label{eq.tr-equivariant}
\end{align}
This system can be seen as $\mathrm{O}(2)$-equivariant \cite{finzi2021practical} (with the gravitational acceleration breaking the full $\mathrm{O}(3)$ symmetry), but we prefer to see it as $\mathrm{O}(3)$-equivariant, with the gravitational acceleration treated as an input to the model.
In particular, if we treat it this way, we might be able to train with one value of $\mathbf{g}$ and generalize at test time to other values of $\mathbf{g}$. 

\section{Learning dynamics from data} 
\begin{figure}[H]
    \centering
    \vspace{-0.25cm}
    \includegraphics[scale=0.25]{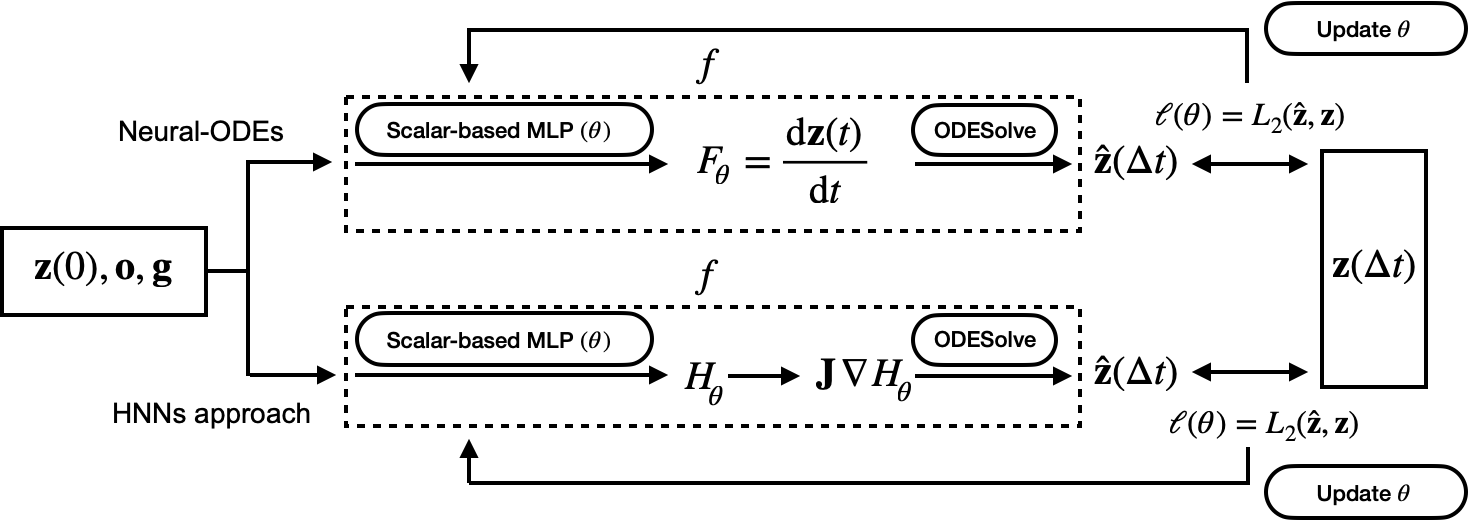} 
    \vspace{-0.3cm}
    \caption{{\small The solving process of learning springy dynamics from data.}}
    \label{fig:process}
\end{figure}

The double pendulum with springs is a Hamiltonian dynamical system, where $\mathcal{H} = \mathcal{T} + \mathcal U$ is time invariant. 
Inspired by \cite{finzi2021practical} we consider two approaches to model $f$ in equation (\ref{eq:goal_F}): Neural ordinary differential equations (N-ODEs) \cite{Chen2018neuralODE}, and Hamiltonian neural networks (HNNs) \cite{greydanus2019hnn,sanchezgonzalez2019hamiltonian}. Both approaches, depicted in Figure \ref{fig:process}, are supervised with the training data described in equation \eqref{eq.training} and loss function $\ell({\theta})=\sum_{i=1}^N\sum_{j=0}^T \Vert \mathbf{z}^i(t_j)-\hat{\mathbf{z}}^i(t_j)\Vert^2$, where $\theta$ represents the parameters in the neural networks to optimize. We implement HNNs and N-ODEs with scalar-based MLPs following the idea from \cite{villar2021scalars} described in Section~\ref{sec:scalars}.

N-ODEs use neural networks to parameterize the derivative function $F(\mathbf{z}(t),t)=\der{\mathbf{z}(t)}{t}$ \cite{Chen2018neuralODE}. Here,
\begin{align}
    \der{\mathbf{v}}{t} = F\big((\q_1,\q_2,\p_1,\p_2), \q_o,\mathbf{g}\big), 
\end{align}
since the dynamics are time-homogeneous. 
The vector function $F$ is $\mathrm{O}(3)$-equivariant with respect to all its inputs, and translation invariant in position. Therefore, there exists a unique function ${F}_{o}$ such that
$F(\mathbf{q}_1,\mathbf{p}_1,\mathbf{q}_2,\mathbf{p}_2,\mathbf{q}_o,\mathbf{g})  ={F}_{o}(\mathbf{q}_1-\mathbf{q}_o,\mathbf{q}_2-\mathbf{q}_o,\mathbf{q}_1-\mathbf{q}_2,\mathbf{p}_1,\mathbf{p}_2,\mathbf{g})$. Note that the input  $\mathbf{q}_1-\mathbf{q}_2$ is redundant, we add it for conceptual reasons, see equation \eqref{eq:energy_U}. We model ${F}_{o}$ as $\mathtt h$ in (\ref{eq:vector_output}) with vector inputs $\mathcal{E}:=\{\mathbf{q}_1-\mathbf{q}_o,\mathbf{q}_2-\mathbf{q}_o,\mathbf{q}_1-\mathbf{q}_2,\mathbf{p}_1,\mathbf{p}_2,\mathbf{g}\}$, obtaining 
\begin{align}
    F_{o}\big((\mathbf{q}_1,\mathbf{p}_1,\mathbf{q}_2,\mathbf{p}_2),\q_o,\mathbf{g}\big) = \sum_{\mathbf{u}\in\mathcal{E}} g_{\mathbf{u}} \big(\{\sigma(\mathbf{e}^\top \mathbf{e}^\prime):\, \forall\, \mathbf{e},\mathbf{e}^\prime\in\mathcal{E},\;\sigma\in\Omega \}\big)\;\mathbf{u},\label{eq:scalar_NODEs}
\end{align}
where $\Omega$ is a set of scalar transformation functions (e.g. $\Omega=\{\sigma_1:x\mapsto x;\;\sigma_2:x\mapsto\sqrt{|x|}\}$). Here the set of functions $\{g_{\mathbf{u}}:\,\mathbf{u}\in\mathcal{E}\}$ are scalar-based and approximated by MLPs. The predicted ``rollout'' trajectory, $\big(\mathbf{z}(0),\hat{\mathbf{z}}(t_1),\ldots,\hat{\mathbf{z}}(t_T)\big)$, is computed iteratively with an ODE solver, 
\begin{align}
   \hat{\mathbf{z}}(t_{j})=\mathrm{ODESolve}\big(\hat{\mathbf{z}}(t_{j-1}),t_{j-1},t_j, F_{o}\big), \quad \hat{\mathbf{z}}(0)=\mathbf{z}(0), \label{eq:ODEsolver_NODEs}
\end{align} 
for each time step $j=1,\ldots,T$.

For the HNN case, we parameterize the Hamiltonian function of the dynamical system with a neural network, and we use a Hamiltonian integrator to predict the dynamics \cite{greydanus2019hnn}. 
The Hamiltonian $\mathcal{H}$ is an invariant scalar function, so similarly to the previous case, we can write $\mathcal{H}(\mathbf{q}_1,\mathbf{q}_2,\mathbf{p}_1,\mathbf{p}_2,\mathbf{q}_o,\mathbf{g}) ={\mathcal{H}}_o(\mathbf{q}_1-\mathbf{q}_o,\mathbf{q}_2-\mathbf{q}_o,\mathbf{q}_1-\mathbf{q}_2,\mathbf{p}_1,\mathbf{p}_2,\mathbf{g})$.
We model $\mathcal{H}$ as $\mathtt f$ in (\ref{eq:scalar_output}), obtaining
\begin{align}
    \mathcal{H}(\mathbf{q}_1,\mathbf{p}_1,\mathbf{q}_2,\mathbf{p}_2,\mathbf{q}_o,\mathbf{g}) = h\big( \{\sigma(\mathbf{e}^\top\mathbf{e}^\prime):\,\forall \,\mathbf{e},\mathbf{e}^\prime \in \mathcal{E},\;\sigma\in\Omega\} \big), \label{eq:scalar_HNNs}
\end{align}
where $\Omega$ is a set of scalar transformation functions as in the N-ODE case, $h$ is a scalar function based on all the inner products of the vectors in $\mathcal{E}$. We approximate $h$ by MLPs. The Hamiltonian dynamics are obtained by integrating the corresponding ODE,
$\der{\mathbf{z}(t)}{t} = \mathbf{J}\nabla \mathcal{H}$, where $\mathbf{J}=[\mathbf{0} \;\;\mathbf{I}\,;\,\mathbf{-I}\;\;\mathbf{0} ]$, with an ODE solver as in equation \eqref{eq:ODEsolver_NODEs} \cite{sanchezgonzalez2019hamiltonian}.


\section{Numerical results}
Our numerical experiments use the same data generation scheme as in \cite{finzi2021practical}. The training data is generated from equation \eqref{eq.training} with $T=5$ and $N=500$. We compare the scalar-based implementation with the constraint-based implementation of the symmetries considered in \cite{finzi2021practical}, i.e., $\mathrm{O}(2)$, the rotation group of order 2 ($\mathrm{SO}(2)$), and the dihedral group of degree 2 and 6 ($\mathrm{D}_2$ and $\mathrm{D}_6$). We also compare to standard non-symmetry enforcing N-ODEs and HNNs models as a baseline. 
The source code is published in \url{https://github.com/weichiyao/ScalarEMLP}.
\begin{figure}[H]
        \centering
        \begin{subfigure}[b]{0.675\textwidth}
            \includegraphics[width=\linewidth]{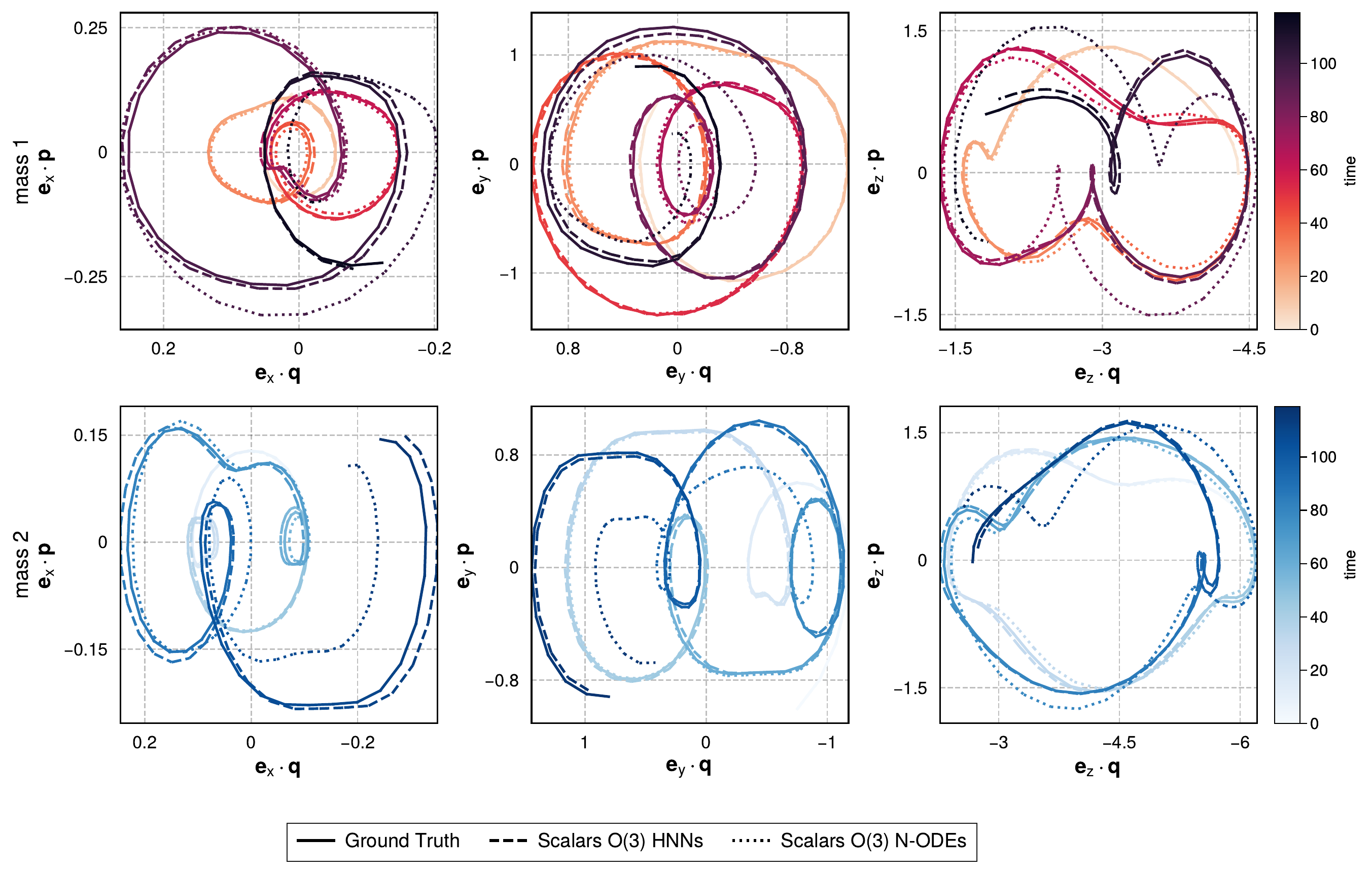} 
        \end{subfigure}%
        \begin{subfigure}[b]{0.325\textwidth}
            \includegraphics[width=\linewidth]{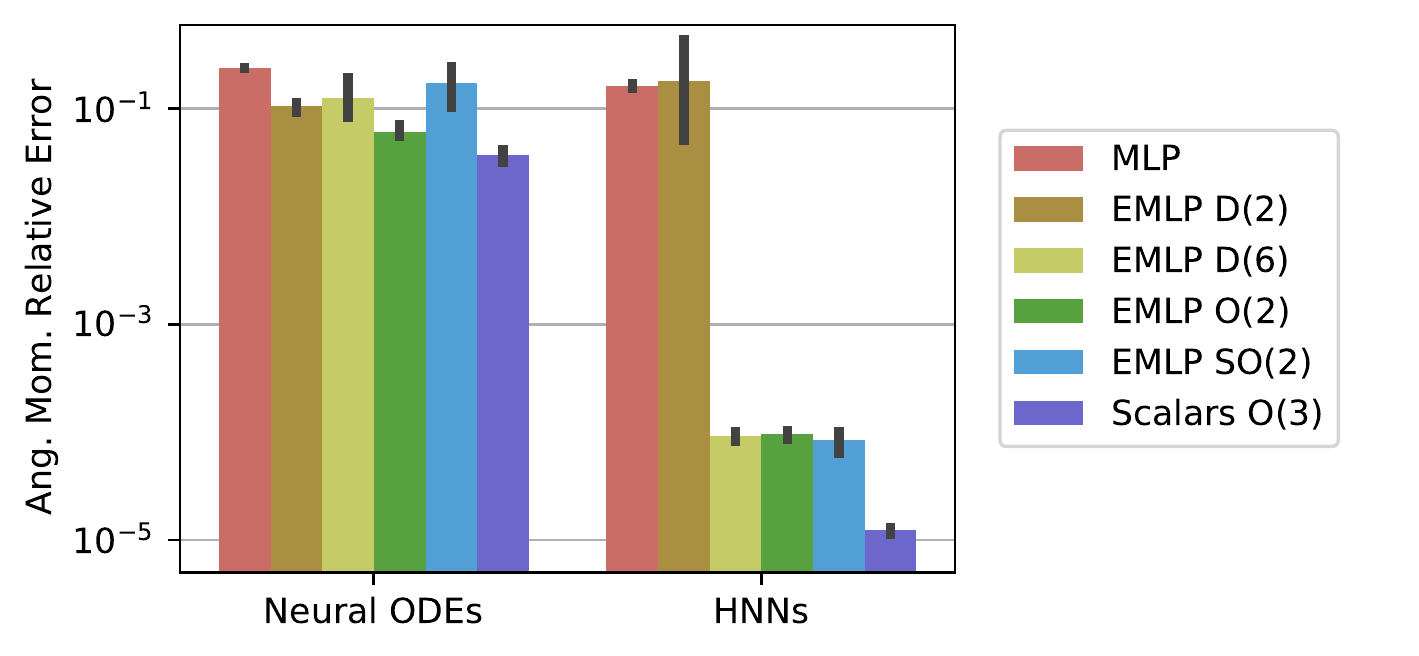}\\
            \vspace{0.25cm}
            \includegraphics[width=\linewidth]{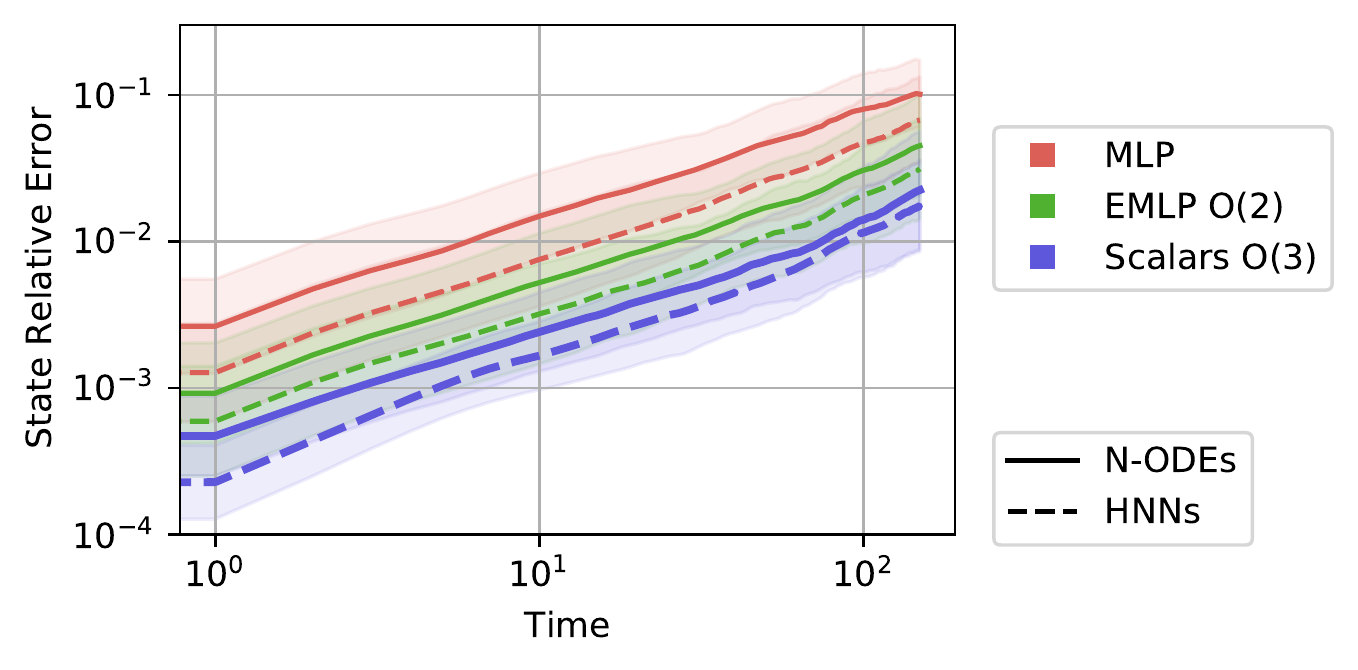}
            \vspace{0.5cm}
        \end{subfigure}%
        \caption{ \small
        \textbf{Left}: Ground truth and predictions of mass 1 (top) and 2 (bottom) in the phase space w.r.t. each dimension. HNNs exhibits more accurate predictions for longer time scales.
        \textbf{Top right}: Invariance of the projection of the angular momentum onto the gravitational force vector $L_\perp$ as in (\ref{eq:ang_momentum}).
        The geometric mean of the relative error is computed over $T=150$ rollouts 
        and averaged across initial conditions. Errorbars are $95$\% confidence interval over 3 runs. 
        \textbf{Bottom right}: The state relative error along the orbit as in (\ref{eq:state_relerr}). 
        Shaded regions show one standard deviation in log space across the different trajectories, showing the variance in the data. Only the best EMLP method is shown for clarity. 
        }
        \label{fig:angular_rollouts}
    \end{figure}

In Figure \ref{fig:angular_rollouts} we show the relative error of the different models in terms of the angular momentum defined as 
\begin{align}
    L_\perp(t)= (\mathbf q_1(t)\times \mathbf p_1(t) + \mathbf q_2(t)\times \mathbf p_2(t))^\top \cdot  \frac{\mathbf g}{\|\mathbf g\|}, \label{eq:ang_momentum}
\end{align}
the projection of the angular momentum onto $\mathbf{g}$. 
We observe, similar to~\cite{finzi2021practical}, that the Hamiltonian-based models conserve the angular momentum as expected, whereas the N-ODEs do not. We also depict the relative state errors, 
defined at a given time $t$ in terms of the positions and momenta of the masses,
\begin{align}
    \text{State.RelErr}(t) =  \frac{\sqrt{(\hat{\mathbf{z}}(t)-\mathbf{z}(t))^\top (\hat{\mathbf{z}}(t)-\mathbf{z}(t))}}{\sqrt{\hat{\mathbf z}(t)^\top\hat{\mathbf z}(t)}+\sqrt{\mathbf z(t)^\top \mathbf z(t)}},\label{eq:state_relerr}
\end{align}
of the scalar-based models along the orbits in Figure \ref{fig:angular_rollouts}. 
Table~\ref{tbl:rollout} gives the geometric mean of these state relative errors over $T=150$. It shows that the scalars approach significantly outperforms the baseline MLP models and the EMLP methods, for both the N-ODE and HNN cases. 

\begin{table}[t!]
\small
    \centering 
    \begin{tabular}{L{1.5cm} C{2cm} C{1.575cm} C{1.575cm} C{1.575cm} C{1.575cm} C{1.575cm} }
    \toprule 
         & \multirow{2}{*}{\textbf{Scalars $\mathrm{O}(3)$}} &\multicolumn{4}{c}{EMLP} &\multirow{2}{*}{MLP}\\
         \cmidrule{3-6}
         &   & $\mathrm{O}(2)$ & $\mathrm{SO}(2)$ & $\mathrm{D}_2$ &  $\mathrm{D}_6$ &\\
         \midrule
         N-ODEs: & $\mathbf{.009\pm .001}$ &$.020\pm .002$& $.051\pm .036$& $.023\pm .002$ & $.036\pm .025$& $.048\pm .000$\\
         HNNs:  & $\mathbf{.005\pm .002}$ & $.012\pm .002$& $.016\pm .003 $& $.111\pm .167$ & $.013\pm .002$& $.028\pm .001$\\
         \bottomrule
    \end{tabular}
     \caption{ {\small Geometric mean (standard deviation computed over 3 trials) of state relative errors of the springy pendulum over $T=150$. Results are shown for the scalar-based MLP HNNs and N-ODEs vs EMLP models from \cite{finzi2021practical} with different symmetry groups, and the standard MLP with no symmetry enforcement. EMLP trains in about one hour, whereas the scalars-based method takes 15 minutes (same machine, same training data). } 
     \vspace{-0.5cm}
     }
    \label{tbl:rollout}
\end{table}

\section{Discussion}

This paper provides a proof of concept implementation for scalar-based equivariant machine learning. Our experimental results for the springy pendulum problem support the intuition that using correct knowledge about the fundamental symmetries of the system leads to better performance on the machine learning algorithms. This not only applies to $\mathrm{O}(3)$ and translation equivariance, but also the Hamiltonian symmetry, as it is apparent in Figure~\ref{fig:angular_rollouts}.

The simplicity of the scalar-based formulation results in fast methods with state of the art performance. However, many technical challenges remain open, for instance establishing the robustness of the approach under noisy data, making the method scalable with respect to the number of input vectors, and investigating the ability of the model to generalize to other settings. One example of the latter is inferring the behavior of the springy pendulum when the masses or natural lengths change. To this end it may be interesting to combine the scalars-based modeling with symbolic regression.

\bibliographystyle{abbrv}
\bibliography{reference}

\begin{thebibliography}{10}

\bibitem{anderson2019cormorant}
B.~Anderson, T.~S. Hy, and R.~Kondor.
\newblock Cormorant: Covariant molecular neural networks.
\newblock In H.~Wallach, H.~Larochelle, A.~Beygelzimer, F.~d\textquotesingle
  Alch\'{e}-Buc, E.~Fox, and R.~Garnett, editors, {\em Neural Information
  Processing Systems}, volume~32. Curran Associates, Inc., 2019.

\bibitem{aniyan2017radio}
A.~K. Aniyan and K.~Thorat.
\newblock Classifying radio galaxies with the convolutional neural network.
\newblock {\em The Astrophysical Journal Supplement Series}, 230(2):20, Jun
  2017.

\bibitem{Chen2018neuralODE}
R.~T.~Q. Chen, Y.~Rubanova, J.~Bettencourt, and D.~K. Duvenaud.
\newblock Neural ordinary differential equations.
\newblock In {\em Advances in Neural Information Processing Systems},
  volume~31. Curran Associates, Inc., 2018.

\bibitem{Chidester2018turbulence}
B.~Chidester, M.~N. Do, and J.~Ma.
\newblock Rotation equivariance and invariance in convolutional neural
  networks.
\newblock {\em arXiv preprint arXiv:1805.12301}, 2018.

\bibitem{choo2018quantum}
K.~Choo, G.~Carleo, N.~Regnault, and T.~Neupert.
\newblock Symmetries and many-body excitations with neural-network quantum
  states.
\newblock {\em Physical Review Letters}, 121:167204, Oct 2018.

\bibitem{cohen2019gauge}
T.~Cohen, M.~Weiler, B.~Kicanaoglu, and M.~Welling.
\newblock Gauge equivariant convolutional networks and the icosahedral cnn.
\newblock In {\em International Conference on Machine Learning}, pages
  1321--1330. PMLR, 2019.

\bibitem{Cohen2016Conv}
T.~S. Cohen and M.~Welling.
\newblock Group equivariant convolutional networks.
\newblock In {\em Proceedings of the 33rd International Conference on
  International Conference on Machine Learning - Volume 48}, page 2990–2999,
  2016.

\bibitem{dominguez-sanchez2018galaxy}
H.~Domínguez~Sánchez, M.~Huertas-Company, M.~Bernardi, D.~Tuccillo, and J.~L.
  Fischer.
\newblock Improving galaxy morphologies for sdss with deep learning.
\newblock {\em Monthly Notices of the Royal Astronomical Society},
  476(3):3661–3676, Feb 2018.

\bibitem{finzi2021practical}
M.~Finzi, M.~Welling, and A.~G. Wilson.
\newblock A practical method for constructing equivariant multilayer
  perceptrons for arbitrary matrix groups.
\newblock {\em arXiv:2104.09459}, 2021.

\bibitem{fuchs2020se}
F.~Fuchs, D.~Worrall, V.~Fischer, and M.~Welling.
\newblock Se (3)-transformers: 3d roto-translation equivariant attention
  networks.
\newblock {\em Neural Information Processing Systems}, 33, 2020.

\bibitem{gonzalez2018augment}
R.~González, R.~Muñoz, and C.~Hernández.
\newblock Galaxy detection and identification using deep learning and data
  augmentation.
\newblock {\em Astronomy and Computing}, 25:103--109, 2018.

\bibitem{greydanus2019hnn}
S.~Greydanus, M.~Dzamba, and J.~Yosinski.
\newblock Hamiltonian neural networks.
\newblock In H.~Wallach, H.~Larochelle, A.~Beygelzimer, F.~d\textquotesingle
  Alch\'{e}-Buc, E.~Fox, and R.~Garnett, editors, {\em Advances in Neural
  Information Processing Systems}, volume~32, 2019.

\bibitem{Gupta_2018}
A.~Gupta, J.~M.~Z. Matilla, D.~Hsu, and Z.~Haiman.
\newblock Non-gaussian information from weak lensing data via deep learning.
\newblock {\em Physical Review D}, 97(10), May 2018.

\bibitem{huertas-company2015galaxy}
M.~Huertas-Company, R.~Gravet, G.~Cabrera-Vives, P.~G. Pérez-González, J.~S.
  Kartaltepe, G.~Barro, M.~Bernardi, S.~Mei, F.~Shankar, P.~Dimauro, and et~al.
\newblock A catalog of visual-like morphologies in the 5 candels fields using
  deep learning.
\newblock {\em The Astrophysical Journal Supplement Series}, 221(1):8, Oct
  2015.

\bibitem{kondor2018n}
R.~Kondor.
\newblock N-body networks: a covariant hierarchical neural network architecture
  for learning atomic potentials.
\newblock {\em arXiv:1803.01588}, 2018.

\bibitem{Ling2016turbulence}
J.~Ling, A.~Kurzawskim, and J.~Templeton.
\newblock Reynolds averaged turbulence modeling using deep neural networks with
  embedded invariance.
\newblock {\em Journal of Fluid Mechanics}, pages 155--166, 2017.

\bibitem{luo2021quantum}
D.~Luo, Z.~Chen, K.~Hu, Z.~Zhao, V.~M. Hur, and B.~K. Clark.
\newblock {Gauge Invariant Autoregressive Neural Networks for Quantum Lattice
  Models}.
\newblock {\em arXiv:2101.07243}, 2021.

\bibitem{Mattheakis2019turbulence}
M.~Mattheakis, P.~Protopapas, D.~Sondak, M.~D. Giovanni, , and E.~Kaxiras.
\newblock Physical symmetries embedded in neural networks.
\newblock {\em arXiv preprint arXiv:1904.08991}, 2019.

\bibitem{sanchezgonzalez2019hamiltonian}
A.~Sanchez-Gonzalez, V.~Bapst, K.~Cranmer, and P.~Battaglia.
\newblock Hamiltonian graph networks with ode integrators, 2019.

\bibitem{schutt2021equivariant}
K.~T. Schütt, O.~T. Unke, and M.~Gastegger.
\newblock Equivariant message passing for the prediction of tensorial
  properties and molecular spectra, 2021.

\bibitem{thomas2018tensor}
N.~Thomas, T.~Smidt, S.~Kearnes, L.~Yang, L.~Li, K.~Kohlhoff, and P.~Riley.
\newblock Tensor field networks: Rotation-and translation-equivariant neural
  networks for 3d point clouds.
\newblock {\em arXiv:1802.08219}, 2018.

\bibitem{vieijra2020quantum}
T.~Vieijra, C.~Casert, J.~Nys, W.~De~Neve, J.~Haegeman, J.~Ryckebusch, and
  F.~Verstraete.
\newblock Restricted boltzmann machines for quantum states with non-abelian or
  anyonic symmetries.
\newblock {\em Physical Review Letters}, 124(9), Mar 2020.

\bibitem{villar2021scalars}
S.~Villar, D.~W. Hogg, K.~Storey{-}Fisher, W.~Yao, and B.~Blum{-}Smith.
\newblock Scalars are universal: \uppercase{E}quivariant machine learning,
  structured like classical physics.
\newblock {\em CoRR}, abs/2106.06610, 2021.

\bibitem{Wang2021dynamics}
R.~Wang, R.~Walters, and R.~Yu.
\newblock Incorporating symmetry into deep dynamics models for improved
  generalization.
\newblock In {\em International Conference on Learning Representations {ICLR}},
  2021.

\bibitem{wang2021molecular}
Z.~Wang, C.~Wang, S.~Zhao, S.~Du, Y.~Xu, B.-L. Gu, and W.~Duan.
\newblock Symmetry-adapted graph neural networks for constructing molecular
  dynamics force fields, 2021.

\bibitem{Weiler2019E2rotation}
M.~Weiler and G.~Cesa.
\newblock General e(2)-equivariant steerable cnns.
\newblock In {\em Advances in Neural Information Processing Systems},
  volume~32, pages 14334--14345, 2019.

\bibitem{weiler2021coordinate}
M.~Weiler, P.~Forr{\'e}, E.~Verlinde, and M.~Welling.
\newblock Coordinate independent convolutional networks--isometry and gauge
  equivariant convolutions on riemannian manifolds.
\newblock {\em arXiv preprint arXiv:2106.06020}, 2021.

\bibitem{Worrall2019scale}
D.~Worrall and M.~Welling.
\newblock Deep scale-spaces: \uppercase{E}quivariance over scale.
\newblock In {\em Advances in Neural Information Processing Systems},
  volume~32, pages 7364--7376, 2019.

\bibitem{Yoo_2018}
J.~Yoo, N.~Grimm, E.~Mitsou, A.~Amara, and A.~Refregier.
\newblock Gauge-invariant formalism of cosmological weak lensing.
\newblock {\em Journal of Cosmology and Astroparticle Physics},
  2018(04):029–029, Apr 2018.

\end{thebibliography}


\ifarxiv%
\fi

\clearpage
\end{document}